\documentclass[conference]{IEEEtran}
\IEEEoverridecommandlockouts
\usepackage{colortbl}
\usepackage{xcolor}
\usepackage{cite}
\usepackage{amsmath,amssymb,amsfonts}
\usepackage{algorithmic}
\usepackage{graphicx}
\usepackage{textcomp}
\usepackage{booktabs}
\usepackage{tabularx}
\usepackage{array,multirow}
\usepackage[caption=false,font=footnotesize]{subfig}
\def\BibTeX{{\rm B\kern-.05em{\sc i\kern-.025em b}\kern-.08em
    T\kern-.1667em\lower.7ex\hbox{E}\kern-.125emX}}
\begin{document}

\title{FRAMES: Failure Recovery And Monitoring of Embodied Skills for Humanoid Loco-Manipulation}

\author{
\IEEEauthorblockN{
Ajay Vikram Periasami\IEEEauthorrefmark{1},
Xinyuan Luo\IEEEauthorrefmark{2},
Haoyu Li\IEEEauthorrefmark{2},
Xianyi Cheng\IEEEauthorrefmark{2}
}

\IEEEauthorblockA{
\IEEEauthorrefmark{1}\textit{Department of Computer Science, Duke University},
Durham, NC, USA \\
\IEEEauthorrefmark{2}\textit{Department of Mechanical Engineering and Materials Science, Duke University},
Durham, NC, USA \\
ajay.vikram@duke.edu,
xinyuan.luo@duke.edu,
haoyu.harry.li@duke.edu,
xianyi.cheng@duke.edu
}
}

\maketitle

\begin{abstract}
Large language model (LLM) planners can decompose natural-language instructions and select reusable robot skills, but choosing the correct skill does not
guarantee successful physical execution. This gap is especially important
in humanoid loco-manipulation, where errors during approach, grasping, transport, or placement can invalidate the remainder of a long-horizon plan. We present FRAMES, a failure-aware supervisory framework for the Unitree G1 humanoid that operates above the CEER whole-body controller \cite{luo2026ceer}. A \emph{Planner Agent} selects subtasks through parameterized mid-level skills,
while a vision-language-model-based \emph{Monitor Agent} evaluates each
skill using temporal multi-view observations and structured robot and contact
evidence. Detected failures stop the active skill and provide grounded
feedback to a \emph{Recovery Agent}. The framework further includes a \emph{Memory Module} for reusing prior skill experience, and geometric grounding via depth and segmentation. We independently evaluate the monitoring module of the framework in MuJoCo using 100 trials comprising 50 failed and 50 successful executions across five tasks. The monitor detects 48 of 50 failures, correctly accepts 46 of 50 successful executions, and achieves 94.0\% overall accuracy. These results provide initial evidence for the monitoring component, while end-to-end evaluation of the complete recovery loop remains ongoing.
\end{abstract}


\section{Introduction}
\label{sec:introduction}
Humanoid robots achieve loco-manipulation by combining locomotion and manipulation to execute long-horizon tasks in human environments. A typical long-horizon task, such as box transport, requires the robot to execute a sequence of subtasks: approaching the object, grasping it, carrying it while walking, and placing it at a new location. The physical coupling of these subtasks means that early errors cascade into downstream failures. An inaccurate approach directly prevents a clean grasp, a premature transition to walking can leave the box unexpectedly supported by the table, and the subsequent motions of walking or turning can ultimately break a secure hold. If these failures go undetected, subsequent actions will be planned using an incorrect estimate of the world state. Achieving reliable humanoid loco-manipulation therefore requires the robot to continuously monitor physical execution, identify failures in real time, and recover autonomously from the observed state.

Recent robotic agents leverage large language models (LLMs) and vision-language models (VLMs) to interpret instructions and select reusable skills. For instance, SayCan ~\cite{ahn2022can} combines language-based planning with learned affordance values, while Inner Monologue ~\cite{huang2022inner} provides execution feedback to an LLM planner. Similarly, CAPE ~\cite{raman2024cape} utilizes failed skill preconditions to generate corrective actions. These approaches primarily use environment feedback to replan between high-level actions, rather than explicitly monitoring the temporal evolution of an ongoing physical skill and interrupting it upon failure.

\begin{figure}[t]
    \centering
    \includegraphics[width=\columnwidth]{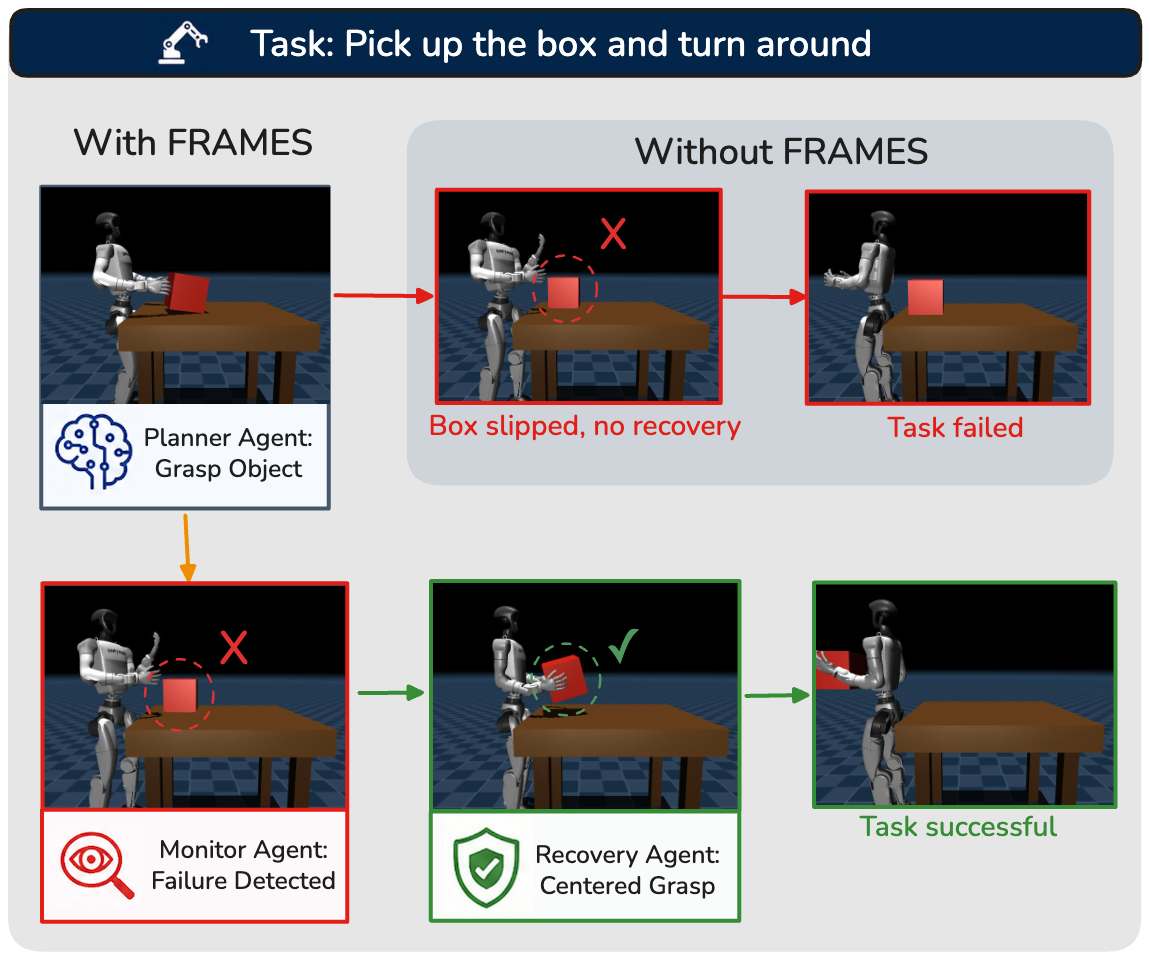}
    \caption{The planner selects the correct grasp subtask, yet the box can still slip during physical execution. Without failure-aware supervision, the robot proceeds from an invalid state and fails the task. FRAMES detects the failed grasp, invokes a centered grasp, and enables the robot to complete the turn while retaining the box.}
    \label{fig:teaser}
\end{figure}

To handle execution errors more robustly, recent methods have moved toward explicit failure reasoning.
REFLECT~\cite{liu2023reflect} and AHA~\cite{duan2024aha} utilize multimodal data and VLMs to detect and explain failures, while RACER~\cite{dai2025racer} applies language feedback to learn recovery behaviors. In the domain of execution monitoring, DoReMi~\cite{guo2024doremi} tracks dynamic constraints via VLMs, Code-as-Monitor~\cite{zhou2025code} synthesizes executable monitoring programs, and COME-robot~\cite{zhi2025closed} integrates visual feedback for mobile-manipulation planning. Despite these advances, most methods focus on fixed-base or wheeled robots, with DoReMi being a notable exception that demonstrates applicability to humanoids. Concurrent to our work, recent humanoid systems like Cybo-Waiter~\cite{ren2026cybo} and POT-VLA~\cite{ren2026closing} have also begun exploring the integration of language planning, geometric grounding, and execution verification. While POT-VLA achieves this by training a unified action model conditioned on shared object tokens, our approach avoids costly policy retraining. Instead, we propose a modular supervisory framework positioned above an existing whole-body controller.

We build our framework on the CEER Controller~\cite{luo2026ceer}, keeping its low-level policy entirely fixed. The core of our contribution is a modular supervisory layer divided into three core entities: a \emph{Planner Agent} to decompose and execute natural-language tasks via mid-level skills, a \emph{Monitor Agent} to check for failures during skill execution, and a \emph{Recovery Agent} to recover from the detected failures. We also develop a \emph{Memory Module} that supplies reusable experience to the \emph{Planner Agent} and \emph{Recovery Agent}. We also provide depth and segmentation skills for the geometric grounding of LLM-suggested skills. This clean separation of concerns enables advanced planning, monitoring, and recovery behaviors without requiring any modifications or retraining at the whole-body control level.

The contributions of this work are:
\begin{itemize}
    \item A modular planning and recovery framework that executes humanoid
    loco-manipulation tasks through parameterized mid-level skills over a fixed
    whole-body controller, with skill parameterization informed by
    persistent subtask-level semantic memory and planner-callable SAM3
    and RGB-D object grounding.

    \item A VLM-based \emph{Monitor Agent} that evaluates each skill using multi-view images, structured robot state, and contact information across a temporal window. It distinguishes ongoing progress from success and failure, stops failed
    skills, and provides failure evidence to the \emph{Recovery Agent}.

    \item A controlled simulation evaluation of the \emph{Monitor Agent}
    over 100 trials spanning failed grasps, dropped objects, collisions
    during approach, wrong-object grasps, and wrong placements. The monitor
    detects 48 of 50 failures, accepts 46 of 50 successful executions,
    achieving 94.0\% overall accuracy.
\end{itemize}

These experiments provide initial quantitative evidence for the monitoring
component. Evaluation of the complete planning, monitoring, recovery, and
memory loop remains ongoing.

\section{Methods}
\label{sec:methods}

\subsection{Problem Formulation}
\label{sec:problem_formulation}

Given a natural-language instruction $g$, the objective is to execute a
sequence of robot skills that satisfies the instruction while detecting and
recovering from execution failures. We use $k$ to index high-level
subtasks, $n$ to index monitoring samples collected during subtask $k$ execution,
and $\ell$ to index low-level CEER control steps. For each subtask, $n=0$
denotes the observation immediately before execution.  The observation at monitoring sample $n$ of subtask $k$ is
\begin{equation}
    o_{k,n}
    =
    \left(
        s_{k,n},
        \mathcal{I}_{k,n}
    \right),
    \qquad
    \mathcal{I}_{k,n}
    =
    \left\{
        I_{k,n}^{v}
    \right\}_{v=1}^{N_{\mathrm{cam}}},
    \label{eq:observation}
\end{equation}
where $s_{k,n}$ is the structured robot and scene state, and
$\mathcal{I}_{k,n}$ contains the available camera views. In simulation,
$N_{\mathrm{cam}}=3$, corresponding to robot-perspective, top-down, and
side-view RGB images. The structured state is
\begin{equation}
\begin{aligned}
    s_{k,n}
    =
    \big(
        &b_{k,n},
        \mathcal{O}_{k,n},
        \mathcal{U}_{k,n}, 
        \mathcal{H}_{k,n},
        \mathcal{K}_{k,n},
        \mathcal{F}_{k,n}
    \big),
\end{aligned}
    \label{eq:structured_state}
\end{equation}
where $b_{k,n}$ is the robot base pose, $\mathcal{O}_{k,n}$ contains
dynamic-object states, $\mathcal{U}_{k,n}$ contains support-surface states,
$\mathcal{H}_{k,n}$ contains hand positions, $\mathcal{K}_{k,n}$ contains
contact relations, and $\mathcal{F}_{k,n}$ contains contact forces.

The instruction is solved sequentially by the \emph{Planner Agent}, one subtask at a time. At
high-level decision step $k$, the system executes the action
\begin{equation}
    a_k
    =
    \left(
        d_k,
        \sigma_k,
        \theta_k,
        C_k
    \right),
    \label{eq:typed_action}
\end{equation}
where $d_k$ describes the intended subtask, $\sigma_k$ is the selected skill
API, $\theta_k$ contains its execution parameters, and $C_k$ is its monitor
contract. For a physical skill, the monitor contract is
\begin{equation}
    C_k
    =
    \left(
        c_k^{\mathrm{pre}},
        c_k^{\mathrm{succ}},
        c_k^{\mathrm{fail}}
    \right),
    \label{eq:monitor_contract}
\end{equation}
where the fields specify the subtask preconditions, success condition, and
failure conditions. Observation and execution-control APIs do not require
a monitor contract.

During each physical skill execution, the \emph{Monitor Agent} periodically
classifies the skill as \texttt{in\_progress}, \texttt{success}, or
\texttt{failure}. An \texttt{in\_progress} verdict allows execution to
continue, while a failure interrupts the skill and activates the
\emph{Recovery Agent}. After the skill finishes, a final monitor check
determines whether control returns to the \emph{Planner Agent}.

\begin{figure*}[t]
    \centering
    \includegraphics[width=\textwidth]{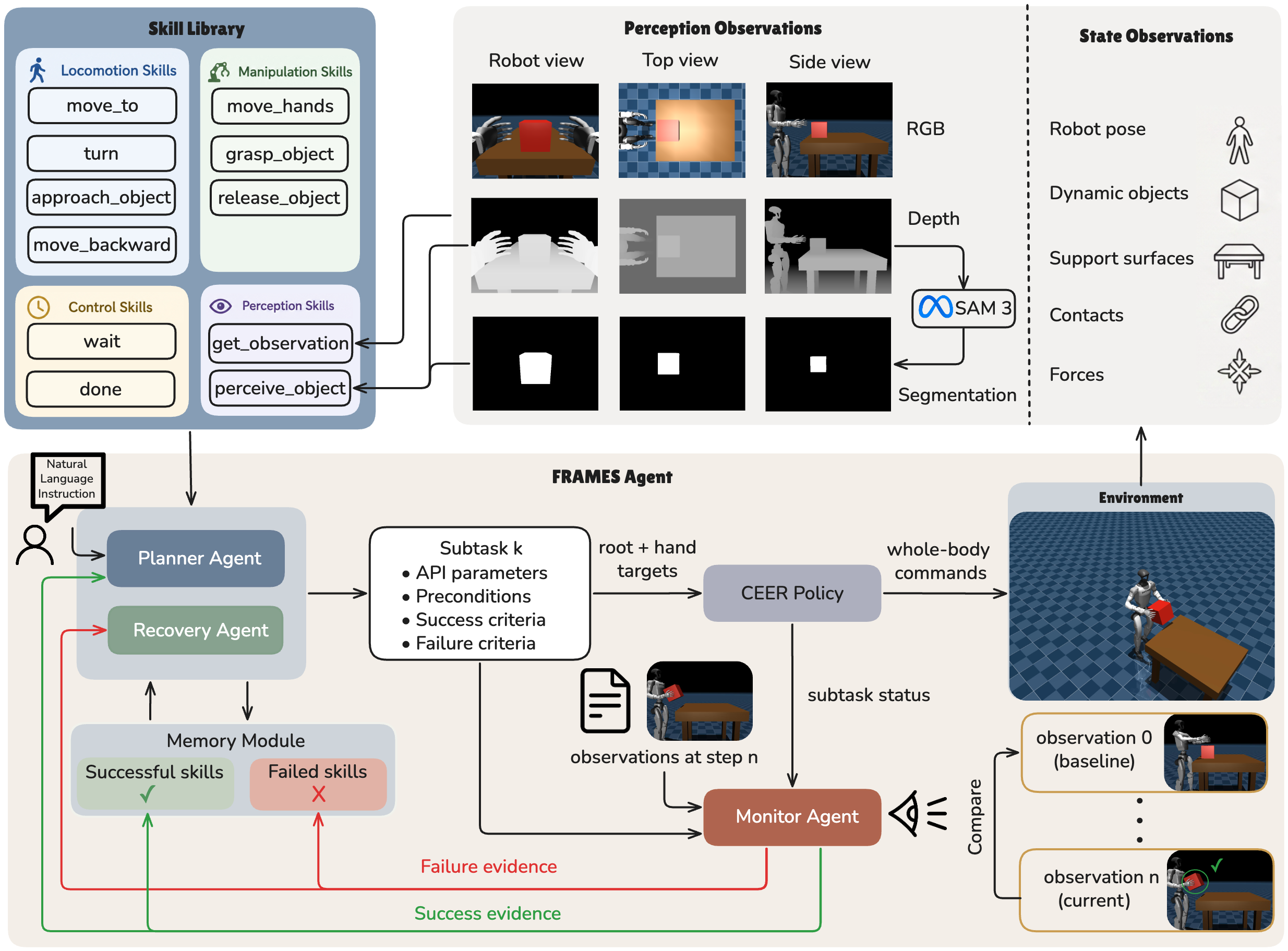}
    \caption{\textbf{Overview of FRAMES.} A natural-language instruction is handled by the \emph{Planner Agent}, which selects a subtask from the skill library together with its API parameters and necessary preconditions and failure and success criteria. The selected subtask is executed through the fixed CEER whole-body controller in the MuJoCo environment. During execution, the \emph{Monitor Agent} compares the current observation against the baseline observation using multi-view perception and structured state information to determine whether the subtask is succeeding or failing. The monitor verdict and supporting evidence are returned immediately to the decision layer, where failures inform the \emph{Recovery Agent}. At the end of each run, completed skill executions are distilled into concise subtask-level records and appended to memory.}
    \label{fig:pipeline}
\end{figure*}

\subsection{Planner Agent}
\label{sec:planner}

At decision step $k$, the \emph{Planner Agent} receives the original
instruction $g$, current observation $o_k \equiv o_{k,0}$, execution
history $h_k$, and skill library $\mathcal{A}$. It proposes one next action:
\begin{equation}
\begin{aligned}
    \tilde{a}_k
    &=
    \left(
        d_k,
        \sigma_k,
        \tilde{\theta}_k,
        C_k
    \right) \\
    &=
    Planner Agent\left(
        g,
        o_k,
        h_k,
        \mathcal{A}
    \right),
\end{aligned}
    \label{eq:planner}
\end{equation}
where $\tilde{\theta}_k$ denotes the initially proposed parameters. The execution history is
\begin{equation}
    h_k
    =
    \left\{
        a_i,
        \rho_i,
        \hat{y}_i,
        z_i
    \right\}_{i=0}^{k-1},
    \label{eq:history}
\end{equation}
where $\rho_i$ is the skill execution result, $\hat{y}_i$ is the \emph{Monitor
Agent} verdict, and $z_i$ is its explanation. Monitor fields are empty for
skills that do not require monitoring. The skill library is
\begin{equation}
    \mathcal{A}
    =
    \mathcal{A}_{\mathrm{nav}}
    \cup
    \mathcal{A}_{\mathrm{man}}
    \cup
    \mathcal{A}_{\mathrm{obs}}
    \cup
    \mathcal{A}_{\mathrm{ctrl}},
    \label{eq:skill_library}
\end{equation}
where the subsets contain navigation, manipulation, observation, and
execution-control skills, respectively. Table~\ref{tab:skill_library} lists
the available skill APIs and their execution arguments. 

\begin{table*}[t]
    \centering
    \caption{Skill Library Available to the Planner and Recovery Agents}
    \label{tab:skill_library}
    \footnotesize
    \setlength{\tabcolsep}{3pt}
    \renewcommand{\arraystretch}{0.96}
    \begin{tabularx}{\textwidth}{
        @{}
        >{\raggedright\arraybackslash}p{0.09\textwidth}
        >{\raggedright\arraybackslash}p{0.15\textwidth}
        >{\raggedright\arraybackslash}p{0.19\textwidth}
        >{\raggedright\arraybackslash}X
        @{}
    }
        \toprule
        \textbf{Category} &
        \textbf{API} &
        \textbf{Arguments} &
        \textbf{Function} \\
        \midrule

        \multirow[t]{4}{*}{Locomotion}
        & \texttt{move\_to}
        & $(x,y,\psi)$
        & Moves the robot base to an absolute world-frame pose. \\

        & \texttt{turn}
        & $(\psi,T)$
        & Rotates the robot toward an absolute world-frame yaw $\psi$ over
          duration $T$. \\

        & \texttt{move\_backward}
        & $(x,y)$
        & Moves the base toward an absolute world-frame position while
          preserving its final orientation. \\

        & \texttt{approach\_object}
        & $(\textit{target},d_{\mathrm{stand}},\Delta\psi)$
        & Moves the base to a target-facing pose with the specified standoff
          distance and optional yaw offset. \\

        \addlinespace[2pt]

        \multirow[t]{3}{*}{Manipulation}
        & \texttt{move\_hands}
        & $(p_L,p_R,T_{\mathrm{hold}})$
        & Moves both hands to robot-base-frame target positions
          $p_L,p_R\in\mathbb{R}^{3}$ and maintains them for
          $T_{\mathrm{hold}}$ seconds, without moving the base. \\

        & \texttt{grasp\_object}
        & $(\textit{target},p_L,p_R,d_{\mathrm{lift}})$
        & Grasps the target using specified left- and right-hand contact
          positions, then lifts it by $d_{\mathrm{lift}}$. \\

        & \texttt{release\_object}
        & None
        & Opens both hands to release the currently held object. \\

        \addlinespace[2pt]

        \multirow[t]{2}{*}{Observation}
        & \texttt{get\_observation}
        & $(\textit{include\_images})$
        & Returns structured state and, when requested, the available camera
          views. \\

        & \texttt{perceive\_object}
        & $(\textit{prompt})$
        & Locates a text-specified object using SAM3 \cite{carion2026sam} and depth, and estimates
          its position, visible size, and hand-to-object distances in the
          robot-base frame. \\

        \addlinespace[2pt]

        \multirow[t]{2}{*}{Control}
        & \texttt{wait}
        & $(T)$
        & Pauses high-level execution for duration $T$. \\

        & \texttt{done}
        & $(r_{\mathrm{claimed}},z_{\mathrm{task}})$
        & Requests task completion with a claimed outcome and summary. \\

        \bottomrule
    \end{tabularx}
\end{table*}

After the Planner Agent selects skill $\sigma_k$ with initial parameters
$\tilde{\theta}_k$, the \emph{Memory Module} retrieves previous records that used
the same skill:
\begin{equation}
    \mathcal{B}_{\sigma_k}
    =
    \left\{
        m_i \in \mathcal{B}
        \mid
        \sigma_i = \sigma_k
    \right\},
    \label{eq:memory_retrieval}
\end{equation}
where $\mathcal{B}$ is the persistent memory bank and
$\mathcal{B}_{\sigma_k}$ contains the retrieved records for the selected
skill.

The same \emph{Planner Agent} is then called in a focused parameterization
mode to reuse prior experience without allowing memory to alter the intended
subtask. It receives the selected skill, initial parameters, current
observation, recent execution history, matching memory records, and the
skill's argument schema. It may revise only schema-permitted execution
parameters, producing $\theta_k$, while $d_k$, $\sigma_k$, the target
object, and $C_k$ remain fixed. If no relevant memory is available or the
revised parameters are invalid, the system retains
$\theta_k=\tilde{\theta}_k$. The resulting action $a_k$, defined in
Eq.~\eqref{eq:typed_action}, is then passed to the \emph{Skill Executor}.

\subsection{Skill Executor: CEER Controller}
\label{sec:skill_execution}

We use the pretrained CEER whole-body controller without modification.
CEER receives root and end-effector targets and maps them, together with
proprioceptive observations, to 29-DoF whole-body commands at 50\,Hz.
Details of its policy architecture, training procedure, and control
abstraction are provided in~\cite{luo2026ceer}.

For a physical skill $\sigma_k$, the \emph{Skill Executor} converts the
selected parameters $\theta_k$ and current state into a sequence of root
and end-effector targets:
\begin{equation}
    u_{k,\ell}^{\mathrm{EE\mbox{-}root}}
    =
    G_{\sigma_k}
    \left(
        \theta_k,
        s_{k,\ell}
    \right),
    \label{eq:skill_controller}
\end{equation}
where $\ell$ indexes low-level control steps. CEER executes these targets as
\begin{equation}
    q_{k,\ell}^{\mathrm{cmd}}
    =
    \pi_{\mathrm{CEER}}
    \left(
        p_{k,\ell},
        u_{k,\ell}^{\mathrm{EE\mbox{-}root}}
    \right),
    \label{eq:ceer_policy}
\end{equation}
where $p_{k,\ell}$ is the CEER proprioceptive observation. The Skill
Executor updates targets at approximately 10\,Hz, while CEER runs at
50\,Hz. 

After a skill completes or is interrupted, the \emph{Skill Executor} returns
\begin{equation}
    \rho_k
    =
    \left(
        r_k^{\mathrm{status}},
        r_k^{\mathrm{message}},
        r_k^{\mathrm{data}}
    \right),
    \label{eq:skill_result}
\end{equation}
containing whether the skill completed and
provides any skill-specific measurements. The \emph{Monitor Agent} separately
evaluates whether the observed execution satisfied the subtask contract
$C_k$. Both outputs are added to the execution history and used to select
the next planning or recovery action.

\subsection{Monitor Agent}
\label{sec:monitor}

The \emph{Monitor Agent} evaluates whether each physical skill satisfies its
subtask contract. It uses skill-local temporal context rather than judging
a single observation independently. Before skill $k$ begins, the framework
captures and retains a baseline observation
\begin{equation}
    o_{k,0}
    =
    \left(
        s_{k,0},
        \mathcal{I}_{k,0}
    \right).
    \label{eq:monitor_baseline}
\end{equation}
At monitoring sample $n$, it captures the current observation
$o_{k,n}$. With three camera views, each monitor query therefore contains
six images: the robot-perspective, top-down, and side views at skill start,
followed by the same three views at the current sample.

The framework also computes a compact structured summary of the change
since skill execution began:
\begin{equation}
    \Delta s_{k,n}
    =
    \phi
    \left(
        s_{k,0},
        s_{k,n}
    \right).
    \label{eq:temporal_delta}
\end{equation}
The summary includes object displacement, change in robot--object distance,
base-pose changes, and transitions in manipulator, support, and floor
contacts. Retaining the same baseline throughout the skill allows the
\emph{Monitor Agent} to evaluate cumulative progress rather than only the change
between adjacent samples. The evidence supplied to the \emph{Monitor Agent} is
\begin{equation}
\begin{aligned}
    e_{k,n}
    =
    \big(
        &g,
        d_k,
        C_k,
        o_{k,0},
        o_{k,n},
        \Delta s_{k,n}
    \big),
\end{aligned}
    \label{eq:monitor_input}
\end{equation}
where $g$ is the complete task instruction, $d_k$ is the current subtask,
and $C_k$ specifies its preconditions, success condition, and failure
conditions. The \emph{Monitor Agent} returns
\begin{equation}
    \left(
        \hat{y}_{k,n},
        z_{k,n}
    \right)
    =
    Monitor Agent\left(e_{k,n}\right),
    \label{eq:monitor_output}
\end{equation}
where $\hat{y}_{k,n}$ is \texttt{in\_progress}, \texttt{success}, or
\texttt{failure}, and $z_{k,n}$ is a concise explanation grounded in the
visual and structured evidence. The \emph{Monitor Agent} checks the skill approximately once per second. An \texttt{in\_progress} result allows execution to continue, while a
\texttt{failure} result stops the skill and activates the \emph{Recovery Agent}.
A \texttt{success} result is accepted only after the skill has finished and
the \emph{Monitor Agent} confirms the final state.

\subsection{Recovery Agent}
\label{sec:recovery}

Let $n_k^{\mathrm{f}}$ denote the monitoring sample at which subtask $k$
is declared failed. The framework constructs the failure record
\begin{equation}
\begin{aligned}
    f_k
    =
    \big(
        &a_k,
        \rho_k,
        z_{k,n_k^{\mathrm{f}}},
        \Delta s_{k,n_k^{\mathrm{f}}},
        o_{k,n_k^{\mathrm{f}}}
    \big).
\end{aligned}
    \label{eq:failure_record}
\end{equation}
The \emph{Recovery Agent} proposes a corrective action:
\begin{equation}
\begin{aligned}
    \tilde{a}_{k+1}^{R}
    =
    Recovery Agent\big(
        &g,
        o_{k,n_k^{\mathrm{f}}},
        h_k,
        f_k,
        \mathcal{A}
    \big).
\end{aligned}
    \label{eq:recovery}
\end{equation}
The \emph{Recovery Agent} uses the same skill library as the \emph{Planner Agent} but is
explicitly conditioned on the failed subtask and its observed failure
evidence.

After the \emph{Recovery Agent} selects a skill, the \emph{Memory Module}
retrieves records for that skill and refines only its permitted execution
parameters just like the \emph{Planner Agent}. The resulting recovery action $a_{k+1}^{R}$ is then executed and monitored using the same procedure as a normal planner action. The \emph{Recovery Agent} may retry the failed skill with revised parameters, select a corrective approach or hand motion, release an unstable contact, or request additional observations.

\subsection{Memory Module}
\label{sec:memory}

The framework maintains a persistent semantic memory bank $\mathcal{B}$.
After each task terminates, including an interrupted run, every completed
physical subtask is distilled into a memory record
\begin{equation}
\begin{aligned}
    m_i
    =
    \big(
        &d_i,
        \sigma_i,
        \theta_i,
        \rho_i,
        r_i^{\mathrm{sub}},
        z_i,
        r_i^{\mathrm{task}},
        \lambda_i
    \big),
\end{aligned}
    \label{eq:memory_record}
\end{equation}
where $r_i^{\mathrm{sub}}$ is the local subtask outcome,
$r_i^{\mathrm{task}}$ is the final whole-task outcome, and $\lambda_i$ is a
concise lesson describing the observed cause, parameter effect, recovery
result, and reuse guidance. Observation and execution-control skills are not
stored.

After either the \emph{Planner Agent} or \emph{Recovery Agent} selects a skill
$\sigma_k$, the Memory Module retrieves records with the same skill:
\begin{equation}
    \mathcal{B}_{\sigma_k}
    =
    \left\{
        m_i \in \mathcal{B}
        \;\middle|\;
        \sigma_i=\sigma_k
        \land
        r_i^{\mathrm{sub}}=r_i^{\mathrm{task}}
    \right\}.
    \label{eq:memory_filter}
\end{equation}
Successful subtasks are retrieved only from successful tasks, while
failed subtasks are retained as negative examples only from failed tasks.
A subtask reported as successful in an ultimately failed task is excluded
because its local success may have been transient or incorrect. Within a
successful task, only the latest successful execution of each skill is
retained.

\subsection{Language-Conditioned Geometric Grounding}
\label{sec:geometric_grounding}

To provide the \emph{Planner Agent} with geometric grounding for more
informed skill parameterization, it can call \texttt{perceive\_object} with
a text description such as ``red box.'' The module applies SAM3 \cite{carion2026sam} to a
frame-matched robot-perspective RGB image to obtain a segmentation mask for
the requested object. Valid depth measurements inside the mask are used to
reconstruct the visible object surface in 3D. Using the camera calibration
and robot base pose, the resulting points are transformed into the world and
robot-base frames. The module returns the visible-surface center, visible
extent, median depth, segmentation confidence, and hand-to-object distances.
These measurements support the parameterization of object-relative
approach, hand-motion, and grasp skills.

\section{Preliminary Experiments \& Results}
\label{sec:results}

\subsection{Experimental Setup}
\label{sec:experimental_setup}

We conduct preliminary experiments in MuJoCo using the 29-DoF Unitree G1
humanoid with the pretrained CEER Controller \cite{luo2026ceer}, which was trained in simulation
using Isaac Lab with 16,384 parallel environments on four NVIDIA L40s GPUs for 8 hours. OpenAI GPT-5.4-mini is used for the \emph{Planner Agent},
\emph{Recovery Agent}, and \emph{Monitor Agent}. SAM3 provides language-conditioned object
segmentation for RGB-D geometric grounding. 

\subsection{Monitor Evaluation Protocol}
\label{sec:monitor_evaluation}

We evaluate the \emph{Monitor Agent} independently on five different tasks. Each task contains 20 trials: 10 failure trials and 10 matched successful executions. Small variations are applied to the robot pose, approach parameters, or manipulation parameters for each of the trials. The required skill sequence is executed by teleoperating, without involving the \emph{Planner Agent} or \emph{Recovery Agent}. This is done to isolate the \emph{Monitor Agent} performance. The five tasks are defined as follows.

\paragraph{Failed grasp.}
The robot is instructed to grasp and lift the box from the table. A trial
is successful when both hands hold and lift the box, and failed when the
box remains on the table after the grasp attempt.

\paragraph{Dropped object.}
The robot begins each trial holding the box above the table and must
maintain its grasp. A trial is successful when the robot continues holding
the box and failed when the box slips or falls onto the table or floor.

\paragraph{Collision during approach.}
The robot is instructed to approach the box without disturbing it. A trial
is successful when the robot reaches the box without contact and failed
when its hands contact and move the box during the approach.

\paragraph{Wrong-object grasp.}
A red box and a blue box of the same shape are placed on the table, and the
robot is instructed to lift the red box. A trial is successful when the
robot lifts the red box and failed when it lifts the blue box.

\paragraph{Wrong placement.}
A blue target pad and a yellow distractor pad are placed on the table, and
the robot is instructed to place the red box on the blue pad. A trial is
successful when the box is placed on the blue pad and failed when it is
placed on the yellow pad. 

We report the number of failures correctly detected, the number of
successful trials correctly accepted and overall classification accuracy. An \texttt{in\_progress} verdict returned after skill completion is counted as an incorrect classification.

\begin{figure*}[!t]
    \centering

    \begin{minipage}[t]{0.188\textwidth}
        \centering
        \includegraphics[width=\linewidth]
        {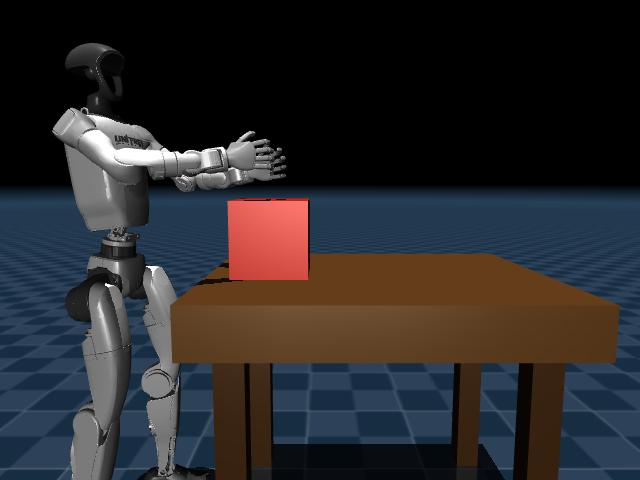}\\[2mm]
        \includegraphics[width=\linewidth]
        {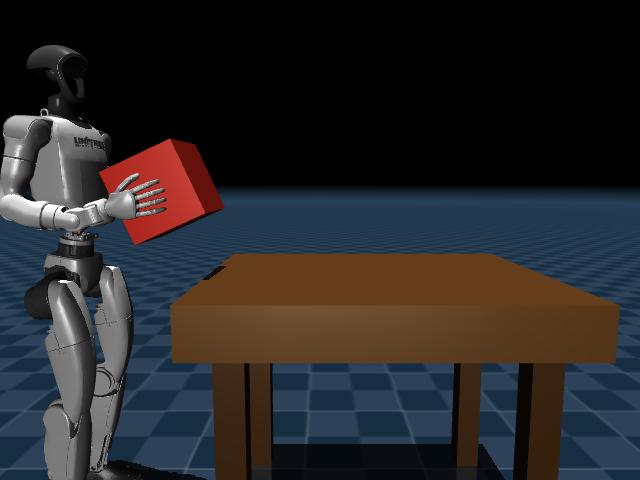}\\[1mm]
        {\footnotesize\textbf{(a)}}
    \end{minipage}
    \hfill
    \begin{minipage}[t]{0.188\textwidth}
        \centering
        \includegraphics[width=\linewidth]
        {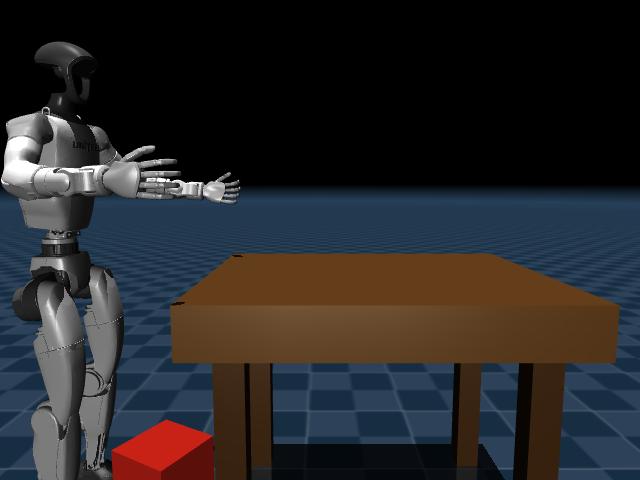}\\[2mm]
        \includegraphics[width=\linewidth]
        {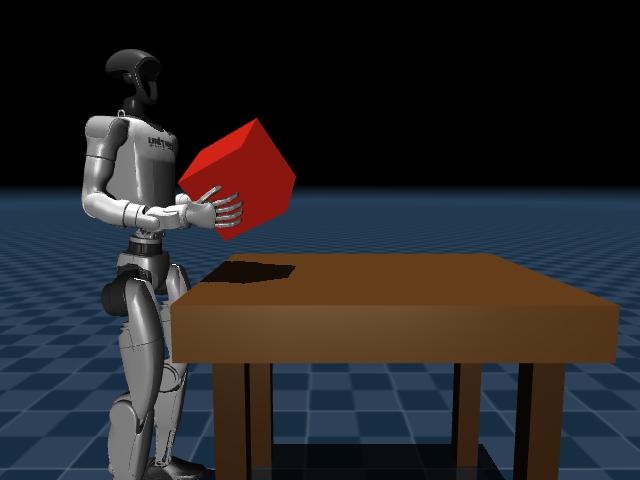}\\[1mm]
        {\footnotesize\textbf{(b)}}
    \end{minipage}
    \hfill
    \begin{minipage}[t]{0.188\textwidth}
        \centering
        \includegraphics[width=\linewidth]
        {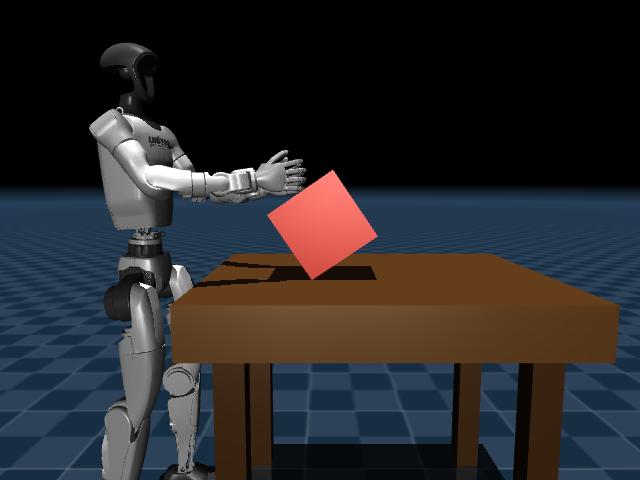}\\[2mm]
        \includegraphics[width=\linewidth]
        {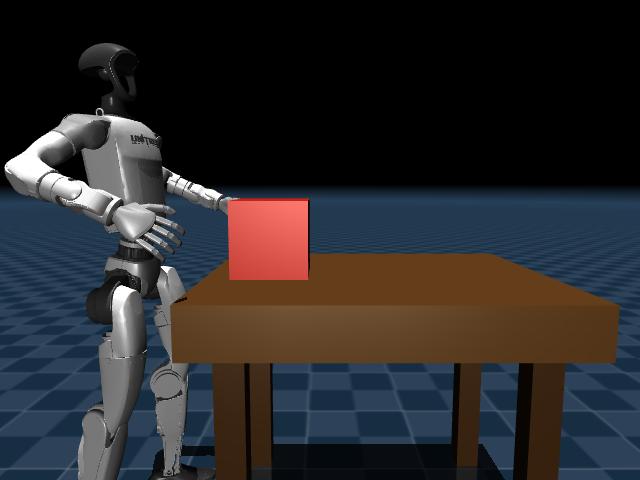}\\[1mm]
        {\footnotesize\textbf{(c)}}
    \end{minipage}
    \hfill
    \begin{minipage}[t]{0.188\textwidth}
        \centering
        \includegraphics[width=\linewidth]
        {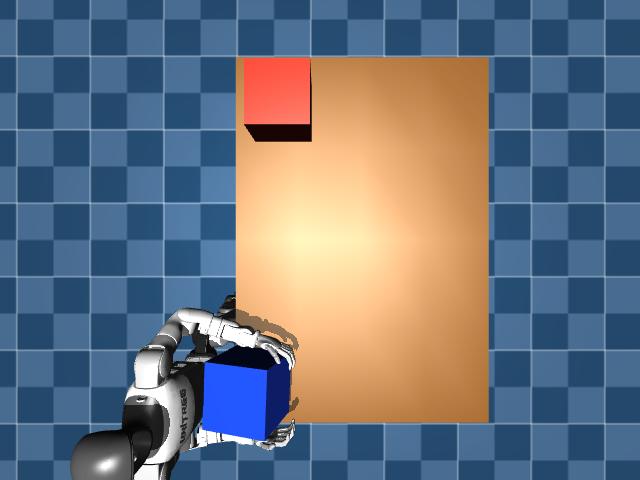}\\[2mm]
        \includegraphics[width=\linewidth]
        {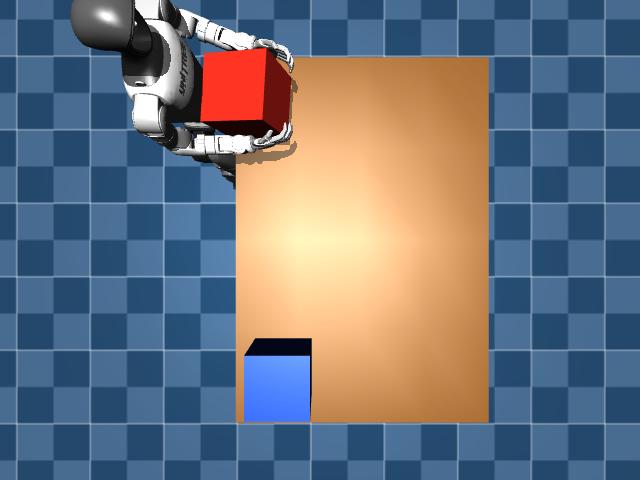}\\[1mm]
        {\footnotesize\textbf{(d)}}
    \end{minipage}
    \hfill
    \begin{minipage}[t]{0.188\textwidth}
        \centering
        \includegraphics[width=\linewidth]
        {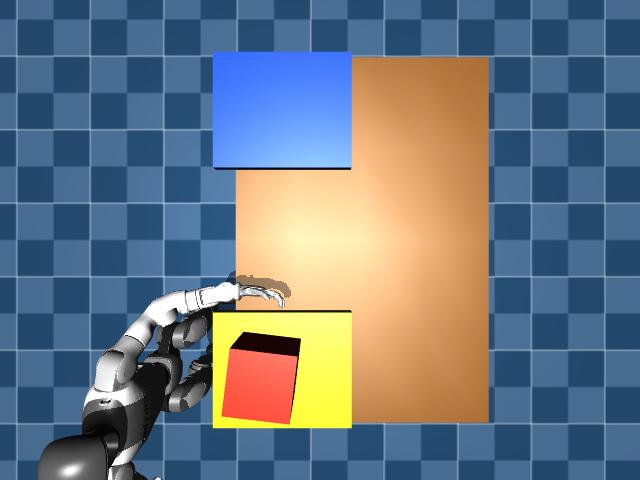}\\[2mm]
        \includegraphics[width=\linewidth]
        {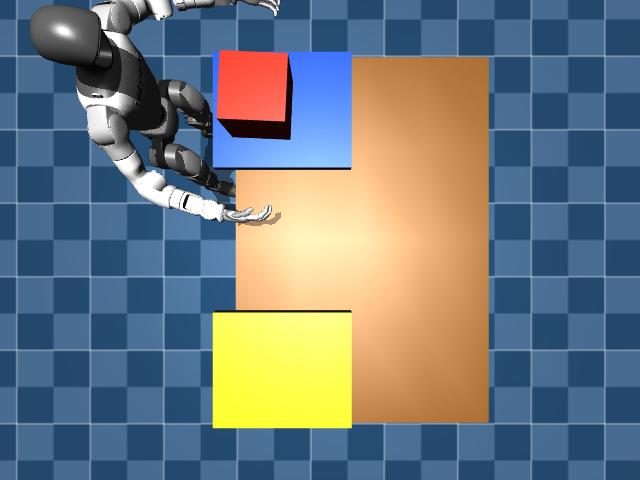}\\[1mm]
        {\footnotesize\textbf{(e)}}
    \end{minipage}

    \caption{Representative failure and success outcomes from the five
    \emph{Monitor Agent} tasks. The top row shows failure conditions, and the
    bottom row shows matched successful executions:
    (a) failed versus successful grasp;
    (b) dropped object versus stable hold;
    (c) collision versus collision-free object approach;
    (d) grasping the blue distractor box versus the requested red box; and
    (e) placing the red box on the yellow distractor pad versus the blue
    target pad.}
    \label{fig:monitor_benchmark_examples}
\end{figure*}

\subsection{Monitor Evaluation Results}
\label{sec:monitor_results}

Table~\ref{tab:monitor_results} summarizes the results. Across all 100
trials, the \emph{Monitor Agent} correctly detects 48 of 50 failures and
accepts 46 of 50 successful executions, corresponding to 94.0\% overall
accuracy. The monitor correctly classifies all dropped-object and wrong-object-grasp trials. The wrong-object task is particularly important because the
monitor identifies the lifted object from camera images without receiving
object names or poses. The remaining errors reveal two limitations. Two failed grasps are labelled
\texttt{in\_progress} even though skill execution has ended. Three safe
approaches are incorrectly classified as collisions because the hands
appear close to the box. One correct placement is also classified as a
failure because the box appears to remain held after release. These errors
show that the monitor is conservative when physical contact is visually
ambiguous. Figure ~\ref{fig:monitor_benchmark_examples} shows qualitative examples of the robot failing and succeeding on the 5 tasks.

\begin{table}[!t]
    \centering
    \caption{Monitor performance across five benchmarks, each containing
    10 failed and 10 successful skill executions.}
    \label{tab:monitor_results}
    \footnotesize
    \setlength{\tabcolsep}{2.5pt}
    \renewcommand{\arraystretch}{1.08}
    \begin{tabular}{@{}lcccc@{}}
        \toprule
        \textbf{Benchmark} &
        \shortstack{\textbf{Failures}\\\textbf{Detected}} &
        \shortstack{\textbf{Successful Runs}\\\textbf{Accepted}} &
        \shortstack{\textbf{Overall}\\\textbf{Accuracy}} \\
        \midrule
        Failed grasp       & 8/10  & 10/10 & 90\% \\
        Dropped object     & 10/10 & 10/10 & 100\% \\
        Collision during approach & 10/10 & 7/10  & 85\% \\
        Wrong-object grasp      & 10/10 & 10/10 & 100\% \\
        Wrong placement    & 10/10 & 9/10  & 95\%  \\
        \midrule
        \textbf{Overall}   &
        \textbf{48/50}     &
        \textbf{46/50}     &
        \textbf{94\%}      \\
        \bottomrule
    \end{tabular}
\end{table}

\section{Conclusion}
\label{sec:conclusion}

We presented FRAMES, a failure-aware framework for humanoid
loco-manipulation built on top of the CEER whole-body controller. The
\emph{Planner Agent} selects parameterized skills and defines how their
execution should be checked. The \emph{Monitor Agent} detects failures using
temporal multi-view images and structured state information, and stops the
current skill when needed. The \emph{Recovery Agent} then selects a corrective
action based on the failure evidence. FRAMES also uses semantic memory to reuse earlier skill experience, and RGB-D and SAM3 to support object grounding.

Our preliminary evaluation focused on the \emph{Monitor Agent}. Across 100
executions from five tasks, it correctly classified 94 trials, detecting
48 of 50 failures and accepting 46 of 50 successful runs. These results show
that the monitoring component can identify several common execution failures,
but the complete planning and recovery loop still requires end-to-end
evaluation. Future work will improve and evaluate the remaining components and test the full framework across a wider range of long-horizon tasks.

\bibliographystyle{IEEEtran}
\bibliography{references}

\end{document}